\documentclass[11pt]{article}
\usepackage{acl2016}
\usepackage{times}
\usepackage{url}
\usepackage{latexsym}
\usepackage{comment}
\usepackage[dvipsnames]{xcolor}
\usepackage{url}
\usepackage{amsmath}
\usepackage{amssymb}
\usepackage{amsfonts}
\usepackage{comment}
\usepackage{verbatim}
\usepackage{graphicx}
\usepackage{caption}
\usepackage{subcaption}
\usepackage{multirow}
\usepackage[font=small,labelfont=bf]{caption}
\usepackage{enumitem}

\usepackage{textcomp}

\DeclareMathOperator{\E}{\mathbb{E}}

\newcommand\todo[1]{\textcolor{blue}{#1}}

\newcommand{\argmax}{\mathop{\rm arg~max}\limits}

\aclfinalcopy 

\title{On the Convergent Properties of Word Embedding Methods}

\author{Yingtao Tian  \qquad Vivek Kulkarni \qquad Bryan Perozzi \qquad Steven Skiena \\
  Stony Brook University \\
  {\tt \{yittian,vvkulkarni,bperozzi,skiena\}@cs.stonybrook.edu}\\\\}

\date{}

\begin{document}
\maketitle

\begin{abstract}
    Do word embeddings converge to learn similar things over different initializations?
	How repeatable are experiments with word embeddings?
	Are all word embedding techniques equally reliable?
	In this paper we propose evaluating methods for learning word representations by their consistency across initializations.	
	 We propose a measure to quantify the similarity of the learned word representations under this setting (where they are subject to different random initializations).
	Our preliminary results illustrate that our metric not only measures a intrinsic property of word embedding methods but also correlates well with other evaluation metrics on downstream tasks.   	
	We believe our methods are is useful in characterizing  \emph{robustness} -- an important property to consider when developing new word embedding methods. 
	\footnote{Copyright \textcopyright 2016 This is the authors draft of the work. It is posted here for your personal use. Not for redistribution.}

\end{abstract}

\section{Introduction}

\begin{table*}[ht!]
	\centering
	\begin{tabular}{||l | c | c | c||}
		\hline
		Category & Question  & \texttt{Glove-1} & \texttt{Glove-2} \\
		\hline
        Syntactics (superlative) &bad : worst $\sim$ slow : ? & \textbf{slowest}         &  pace \\
        Syntactics (opposite) & acceptable : unacceptable $\sim$ ethical : ? & moral & \textbf{unethical}\\
		Semantics (capital\&state) & London : England $\sim$ Madrid : ? & \textbf{Spain} &       Ronaldo\\
		Semantics (family) & brothers : sisters $\sim$ grandson : ?  &  niece  & \textbf{granddaughter} \\
		\hline
	\end{tabular}
	\caption{
	   \small
	    A set of examples of questions in word analogy task \cite{mikolov2013exploiting} where two
	    independent runs of GloVe embeddings differ in the answers.
	    Two questions are picked from the syntactic section and two from the semantic section of the task.
	    The correct answer is highlighted in bold.
	}
	\label{tab:questions}
\end{table*}

Word embeddings learned using neural methods have been shown to be tremendously effective on several NLP tasks \cite{mikolov2010recurrent,chen2014fast,socher2013recursive,2016arXiv160405529P}. 
But the question \emph{``What properties are exhibited by word embeddings that are learned by different techniques?''} remains largely unexplored. 
\newcite{mikolov2013linguistic} show that embeddings learned by Skip-Gram model reveal linear substructures. 
\newcite{arora2015random} attempt to explain this linear substructure by proposing a generative model based on random walks over discourse vectors and \cite{arora2016linear} show that the different senses of a word can be recovered by a sparse coding in such embedding spaces. 

In this work, we carry forward this line of research to understand the different properties of word embedding methods. 
First we motivate our research direction by noting the following: 
(a) The optimization to learn word embeddings is usually non-convex and hence depends on the random initialization and 
(b) Word embeddings (like skip-gram or GloVe) learned by neural models under different random initializations mostly converge to yield very similar performances on NLP tasks. 
This observation begs the following question:
\emph{Do word embeddings obtained under two different random initializations actually learn equivalent features?   
If not, is their similar performance on downstream tasks merely a coincidence?} 
To illustrate with an example, consider the performance of GloVe embeddings obtained on two independent runs (differing only in the random initialization) on the analogy task \cite{mikolov2013exploiting} in Table \ref{tab:questions}. 
Although both embeddings show a high accuracy on this task (see Table \ref{tab:acc}), 
a closer look at the questions where they disagree reveals differences in the relationships they learn. 
An improved understanding of such properties of word embedding methods will enable the development of improved models and learning algorithms (which will will in-turn boost performance on downstream NLP tasks).

Specifically our contributions are as follows:

\noindent
\begin{itemize}
\item \textbf{Mapping-based Similarity Measure between Dimensions}:
We propose a similarity measure between dimensions of two sets of embeddings,
based on correlations using one-to-one or many-to-one linear transformation between these dimensions.
\item \textbf{Intrinsic Way to Evaluate Embeddings}:
Based on this similarity measure, 
we propose a new, intrinsic way to quantify the similarity between two such embedding spaces.  Our preliminary results show that our measure correlates with a measure of model agreement on the well known analogy task.

\end{itemize}

\section{Background and Setup}
\label{sec:setup}

\begin{table}[t!]
	\centering
	\begin{tabular}{||c | c c c | c||}
		\hline
		Emb.  & Sem. & Syn. & Tot. & $\alpha$ \\
		\hline
		\texttt{GloVe-1} & $79.75$ & $69.28$ & $74.02$ & \multirow{2}{*}{$0.89$} \\
		\texttt{GloVe-2} & $79.67$ & $69.34$ & $74.01$ &  \\
		\hline
		\texttt{SG-1} & $67.72$ & $64.34$ & $65.83$ & \multirow{2}{*}{$0.79$} \\
		\texttt{SG-2} & $68.37$ & $63.62$ & $65.72$ & \\
		\hline
	\end{tabular}
	\caption{
    \small
	Performance of two independent runs of GloVe and skip-gram on word analogy task \cite{mikolov2013exploiting}.
	$\alpha$ represents the agreement on the answers between the two runs as computed by 
	Krippendorff's alpha.}
	\label{tab:acc}
\end{table}

Here we describe the datasets and word embeddings we use in our experiments.
We analyze two word embeddings \textbf{GloVe} \cite{pennington2014glove} and \textbf{skip-gram (SG)} \cite{mikolov2013exploiting} where we set the number of dimensions $D$ to be $300$. 
Our corpora consists of a combination of English Wikipedia dump with $1.6$ billions tokens and the English section of the News Crawl\footnote{Available at \url{http://www.statmt.org/wmt16/translation-task.html}. 
We use articles from 2007 to 2015.} with $4.3$ billion tokens totaling to $6$ billion tokens. 
We use similar settings as \newcite{pennington2014glove}. We set the vocabulary to the $400,000$ most frequent words,  
use a context window of size $10$, and make $50$ iterations through the whole dataset. 
We use News Crawl for two reasons: It is a public available corpus of the same genre as Gigawords 5 (news in formal English), and News Crawl is also available in Czech, German, Finish and Russia which allows our analysis to extend to multi-lingual settings.

We train both Glove and skip-gram embeddings twice using different random initializations and order of corpus, and refer to them as \texttt{GloVe-1}, \texttt{GloVe-2}, \texttt{SG-1}, \texttt{SG-2} respectively.
The accuracies of these embeddings on the word analogy task is shown in Table \ref{tab:acc}.
Note that these accuracies are comparable to scores obtained on the same task in \newcite{pennington2014glove}.
We denote these $4$ embeddings as 
$\mathbf{E}^{(m)}$ (a matrix of dimensions $|V| \times D$)
where $m \in \{ \texttt{GloVe-1}, \texttt{GloVe-2}, \texttt{SG-1}, \texttt{SG-2} \}$.

Let $\mathbf{E}_{i}^{(m)}$ denote feature dimension $i$ of $\mathbf{E}^{(m)}$ (column $i$ of $\mathbf{E}^{(m)}$). We now seek to quantify the similarity between embeddings $\mathbf{E}^{(m)}$ and $\mathbf{E}^{(n)}$, the methods for which we discuss in forthcoming sections.

\begin{table}[h!]
\centering

\end{table}

\section{Similarities of Word Embeddings}
\label{sec:similarity}

\begin{figure*}[ht]
  \centering
  
  \begin{subfigure}{.45\textwidth}
  	\centering
    \includegraphics[width=\linewidth]{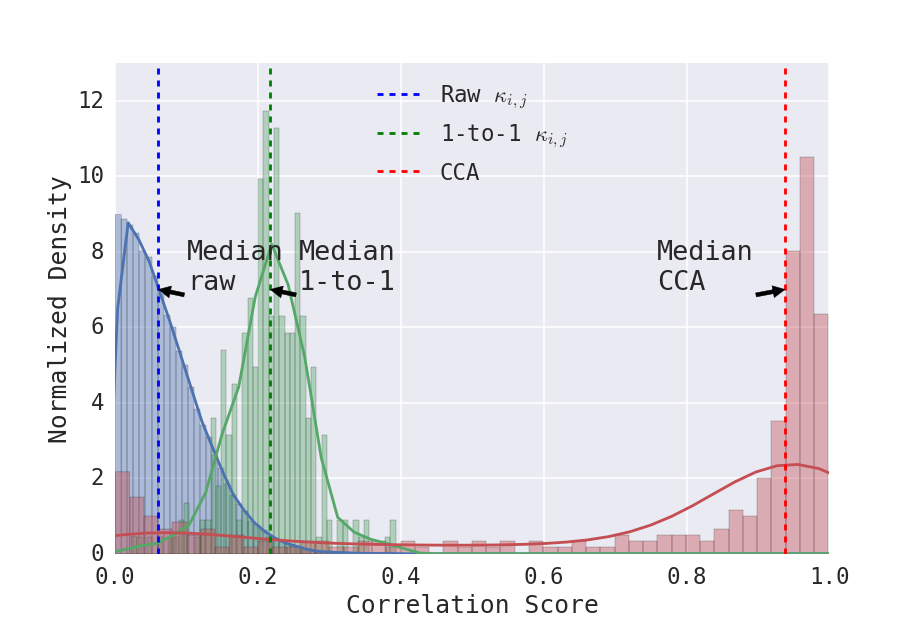}  
    \caption{Histogram of feature-wise correlations between\\ two runs of GloVe.}
    \label{fig:hist-gv}
  \end{subfigure}%
  \begin{subfigure}{.45\textwidth}
    \includegraphics[width=\linewidth]{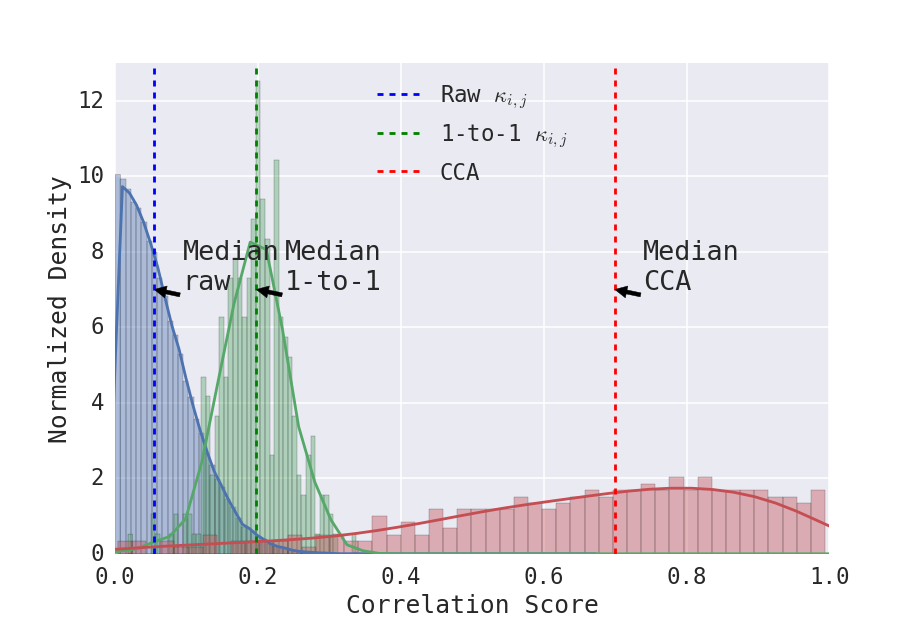}  
    \caption{Histogram of feature-wise correlations between\\ two runs of skip-gram.}
    \label{fig:hist-sg}
  	\centering
  \end{subfigure}
    
  \caption{
    \small
    \textbf{Distributions of Correlations} -- 
    Histogram of 
    {correlations of all raw values in $\kappa (i,j)$ in blue},
    {correlations of 1-to-1 matched  $\kappa (i,j)$ in green},
    and 
    {sample canonical correlations from CCA in red},
    discussed in Section \ref{sec:measure-of-similarity}, \ref{sec:one-to-one}
    and \ref{sec:many-to-one}, respectively. 
    Results between two runs of GloVe embeddings are shown in 
    Subfigure \ref{fig:hist-gv}, 
    and results between two runs of skip-gram embeddings are shown in 
    Subfigure \ref{fig:hist-sg}.
    Also shown are kernel density estimates and medians of corresponding correlations.
  }
  \label{fig:hist}
\end{figure*}
  
\begin{figure}[t!]
	\vspace{-0.2in}
  \includegraphics[width=\linewidth]{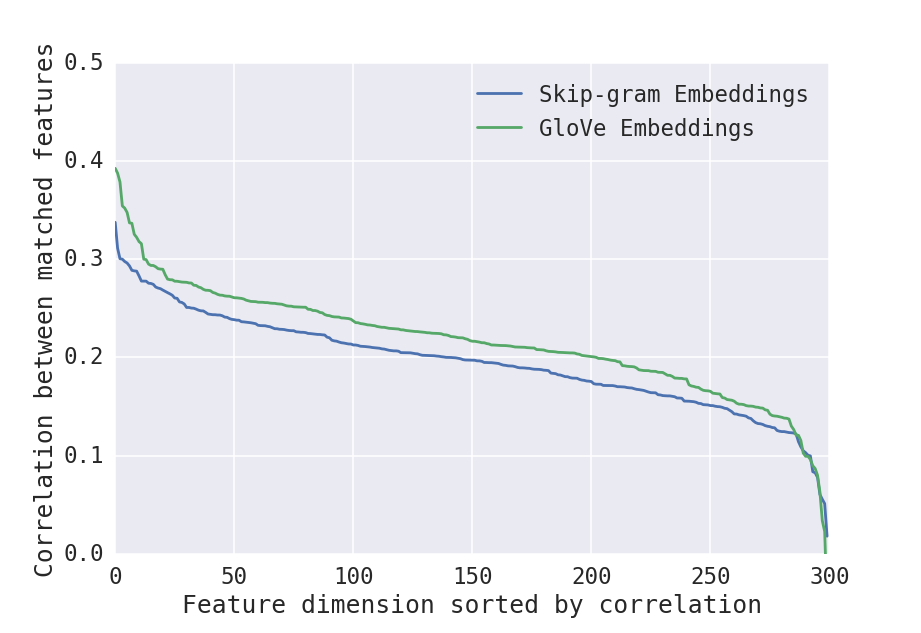}  
  \caption{
  \small
  Correlations between corresponding features obtained using one-to-one matching,
  sorted in descending order.
  Both GloVe and skip-gram demonstrate moderate correlation. 
  GloVe is marginally better than skip-gram.
  Note that random embeddings demonstrate correlations of $0$.
  }
  \label{fig:1to1-dist}	
\end{figure}

\noindent
How similar are the embeddings learned from independent runs over the same corpus?
More formally, given a pair of $D$ dimensional embeddings $\mathbf{E}^{(m)}$ and $\mathbf{E}^{(n)}$ differing only in their initial random initialization:
\begin{itemize}[noitemsep,topsep=0pt]
  \setlength{\itemsep}{1pt}
  \setlength{\parskip}{0pt}
  \setlength{\parsep}{0pt}
\item Can we define a measure of similarity between the dimensions ($\mathbf{E}^{(m)}_{i}$, $\mathbf{E}^{(n)}_{j}$)?
\item Given such a measure, can we align dimensions in $\mathbf{E}^{(m)}$ to dimensions in $\mathbf{E}^{(n)}$ to quantify the similarity between $\mathbf{E}^{(m)}$ and $\mathbf{E}^{(n)}$?
\end{itemize}

\subsection{Measure of Similarity between Feature Dimensions}
\label{sec:measure-of-similarity}
Given $\mathbf{E}^{(m)}$ and $\mathbf{E}^{(n)}$ we define the similarity between 
them 
as $\kappa(i, j)=\rho(\mathbf{E}^{(m)}_{i},\mathbf{E}^{(n)}_{j})$ where $\rho(\textbf{x}, \textbf{y})$ is defined as the Pearson Correlation Coefficient between the column vectors \textbf{x} and \textbf{y}.
We note that this similarity measure is well-suited to capture linear relationships between feature dimensions. 
We leave exploring metrics that capture non-linear relationships (like mutual information) between feature dimensions to future work.

In Figure \ref{fig:hist} we show in blue, the histogram of values in correlation matrices $\kappa(i,j)$ on $\mathbf{E}^{(m)}$ and $\mathbf{E}^{(n)}$ obtained on the Glove and skip-gram embeddings.

\subsection{One-to-one Alignment}
\label{sec:one-to-one}
Since the optimization problems involved in GloVe and skip-gram are inherently non-convex,
the embeddings learned could potentially correspond to different local optima. 
This implies the features learned on different runs could be equivalent under some manifold transformation. 
Therefore as a first step, we ask: Is there a one-to-one alignment between features in two embeddings? (similar to \newcite{li2016ICLR})

Therefore, given $\mathbf{E}^{(m)}$ and $\mathbf{E}^{(n)}$, we seek to find an one-to-one matching
represented as $\mathbf{a} = (a_1, a_2, \dots a_D)$
meaning that for each $d \in \{1, 2, \dots, D\}$,
$\mathbf{E}_{d}^{(n)}$ is matched with $\mathbf{E}_{a_d}^{(m)}$,
where the aggregated correlation is given by
\begin{equation}
\zeta_{\texttt{1to1}} = \frac{1}{D} \sum_{d=1}^{D} \rho(\mathbf{E}_{d}^{(n)}, \mathbf{E}_{a_d}^{(m)}).
\end{equation}
By building a complete bipartite graph,
this can be converted into an instance of \emph{maximum weighted bipartite matching} \cite{douglas1996west},
which can be solved effectively using Hopcroft-Karp Algorithm \cite{hopcroft1973n} in polynomial time. 

This perfect bipartite matching allows us to permute features (columns) of $\mathbf{E}^{(m)}$ to correspondingly match features in  $\mathbf{E}^{(n)}$.
In Figure \ref{fig:hist}, we show in green the histogram of matched correlations 
for both GloVe (Figure \ref{fig:hist-gv}) and skip-gram (Figure \ref{fig:hist-sg}).

Finally, we show the correlations between matched dimensions 
in Figure \ref{fig:1to1-dist}. 
Observe that using the one-to-one alignment between the dimensions of the embeddings, 
the Glove embeddings display a stronger correlation ($21.6\%$) than skip-gram ($19.5\%$). 
Note that random embeddings show correlations of $0$.
We believe this indicates that Glove embeddings are more stable and consistent than skip-gram under random initialization.

\subsection{Many-to-one Mapping}

\label{sec:many-to-one}

In previous section we sought a one-to-one alignment between the features in two embeddings,
One drawback of which 
is its restrictiveness 
since it assumes both models learn exactly the same set of features equivalent under permutations. 
In this section, we hypothesize that a feature in $\mathbf{E}^{(m)}$ could correspond to multiple features in $\mathbf{E}^{(n)}$. 
Hence we relax the restrictions 
to a many-to-one matching,
by assuming a single dimension in $\mathbf{E}^{(n)}$ 
can be obtained to some degree from a linear transformation of multiple dimensions in $\mathbf{E}^{(m)}$ .
We capture this by measuring linear relationship between 
$\textbf{E}^{(n)}$ and $\textbf{E}^{(m)}$ transformed using CCA (Canonical Correlation Analysis).

Given two matrices $X \in \mathbb{R}^{|T|\times d_1}$ and 
			$Y\in \mathbb{R}^{|T|\times d_2}$, 
CCA finds two projections $P \in \mathbb{R}^{|T|\times d}$ and $Q \in \mathbb{R}^{|T|\times d}$
($d \leq \min(d_1, d_2)$)
that maximizes the correlation $\rho(P\boldsymbol{x_i}, Q \boldsymbol{y_j})$ between corresponding columns
\begin{equation}
\argmax_{P,Q}  \sum_{i=1}^{d} \rho(P\boldsymbol{x_i}, Q\boldsymbol{y_i}).
\end{equation}
Using the projection matrices $P$ and $Q$,
 one can now project the inputs $X$ and $Y$ to yield $U = PX$  and $V = QY$ such that 
$\rho(P\boldsymbol{x_i}, Q\boldsymbol{y_i})$ is maximized.
Therefore we propose to use CCA on two embeddings $\textbf{E}^{(n)}$ and $\textbf{E}^{(m)}$ of dimension $D$ and define the similarity between $\textbf{E}^{(n)}$ and $\textbf{E}^{(m)}$ as
\begin{equation}
\zeta_{\texttt{CCA}} = \frac{1}{D} \sum_{i=1}^{D} \rho(P\boldsymbol{x_i}, Q\boldsymbol{y_i}).
\end{equation}

\begin{figure}[h]
  \centering
  \includegraphics[width=\linewidth]{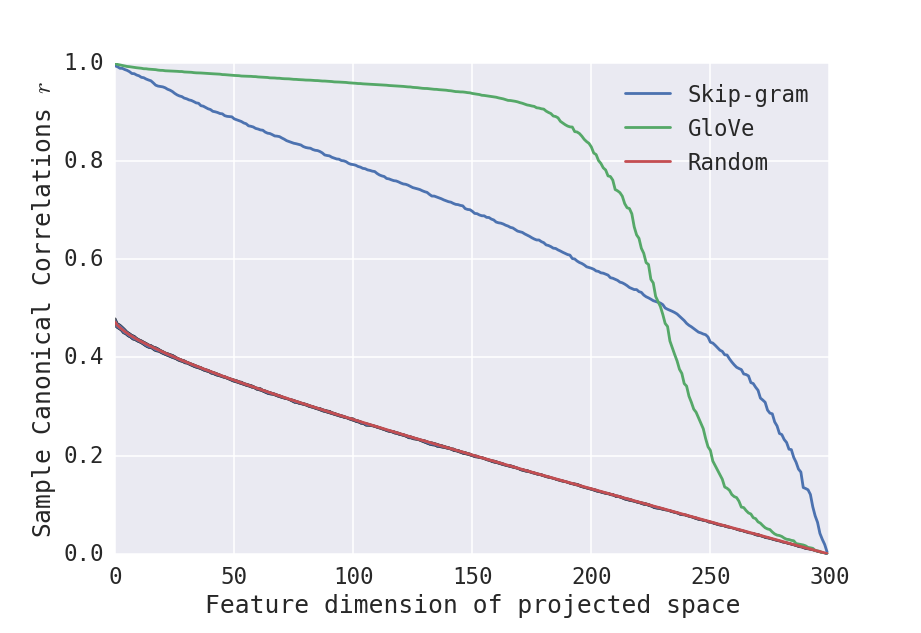}
  \caption{
  \small
  Sample canonical correlations obtained from CCA.
  Note that CCA by design gives dimensions in descending order of sample canonical correlations.
  GloVe demonstrates near perfect correlations in most correlated $200$ dimensions.
  Aggregatively GloVe is better than skip-gram measured by $\zeta_{\texttt{CCA}}$.
  We also show the performance of random embeddings in red.
  }
  \label{fig:cca}
\end{figure}

In Figure \ref{fig:hist}, we show in red,
the histogram of sample canonical correlations obtained from CCA for GloVe and skip-gram embedding methods.
These values stand out from one-to-one matching of features,
and show different distributions for GloVe and skip-gram.
Also in Figure \ref{fig:cca} we show the sample canonical correlations obtained in descending order.
Both skip-gram and GloVe demonstrate strong correlations on several dimensions as compared to random embeddings.
It is also clear that GloVe is better than skip-gram as measured by $\zeta_{\texttt{CCA}}$.
This is consistent with previous observation in Figure \ref{fig:1to1-dist}.

Finally, we show that our proposed metric, $\zeta_{\texttt{CCA}}$,
correlates well with the agreement on the analogy task,
as shown in Table \ref{tab:measure}.
We emphasize that our metric is completely \emph{intrinsic}, 
does not require any external data for evaluation, 
and yet captures useful convergent properties of embedding models.

\begin{table}[h]
	\centering
	\begin{tabular}{||c | c | c ||}
		\hline
		Emb.   & $\alpha$ & $\zeta_{\texttt{CCA}}$\\
		\hline
		\texttt{GloVe-1} and \texttt{GloVe-2}  & {$0.89$}  & {$0.74$} \\
		\hline
		\texttt{SG-1} and \texttt{SG-2} & $0.79$  & $0.66$\\		\hline
	\end{tabular}
	\caption{
	\small
	$\zeta_{\texttt{CCA}}$ correlates well with the agreement scores (represented by $\alpha$ computed by Krippendorff’s alpha) on the analogy task.
	Note that higher $\zeta_{\texttt{CCA}}$ implies higher $\alpha$.
	}
	\label{tab:measure}
\end{table}

\section{Conclusion}

In this paper we proposed an intrinsic metric for evaluating word embeddings --  their consistency across random initializations.
Our preliminary results showed that while total performance 
remained consistent across embeddings, there was a sizeable disagreement between each instance of a particular embedding.
To examine this difference in more depth, we investigated the similarity between the dimensions of each embedding space.  
Furthermore we propose methods to align dimensions of word embeddings 
and an intrinsic metric $\zeta_{cca}$ to measure the similarity of the learned word embeddings,
which our preliminary results show may correlate with downstream tasks in terms of model agreement,
with the popular word analogy task in \newcite{mikolov2013exploiting} as an example. 
We believe the methods and metrics proposed in our work will enable a deeper investigation into the convergent properties of embedding models, 
and  lead to improved optimization algorithms and performance on downstream NLP tasks. 


\bibliography{emb-eval}

\begin{thebibliography}{}

\bibitem[\protect\citename{Arora \bgroup et al.\egroup }2015]{arora2015random}
Sanjeev Arora, Yuanzhi Li, Yingyu Liang, Tengyu Ma, and Andrej Risteski.
\newblock 2015.
\newblock Random walks on context spaces: Towards an explanation of the
  mysteries of semantic word embeddings.
\newblock {\em arXiv preprint arXiv:1502.03520}.

\bibitem[\protect\citename{Arora \bgroup et al.\egroup }2016]{arora2016linear}
Sanjeev Arora, Yuanzhi Li, Yingyu Liang, Tengyu Ma, and Andrej Risteski.
\newblock 2016.
\newblock Linear algebraic structure of word senses, with applications to
  polysemy.
\newblock {\em arXiv preprint arXiv:1601.03764}.

\bibitem[\protect\citename{Chen and Manning}2014]{chen2014fast}
Danqi Chen and Christopher~D Manning.
\newblock 2014.
\newblock A fast and accurate dependency parser using neural networks.
\newblock In {\em Empirical Methods in Natural Language Processing (EMNLP)}.

\bibitem[\protect\citename{Hopcroft and Karp}1973]{hopcroft1973n}
John~E Hopcroft and Richard~M Karp.
\newblock 1973.
\newblock An n\^{}5/2 algorithm for maximum matchings in bipartite graphs.
\newblock {\em SIAM Journal on computing}, 2(4):225--231.

\bibitem[\protect\citename{Li \bgroup et al.\egroup }2016]{li2016ICLR}
Yixuan Li, Jason Yosinski, Jeff Clune, Hod Lipson, and John Hopcroft.
\newblock 2016.
\newblock Convergent learning: Do different neural networks learn the same
  representations?
\newblock In {\em International Conference on Learning Representation (ICLR
  '16)}.

\bibitem[\protect\citename{Mikolov \bgroup et al.\egroup
  }2010]{mikolov2010recurrent}
Tomas Mikolov, Martin Karafi{\'a}t, Lukas Burget, Jan Cernock{\`y}, and Sanjeev
  Khudanpur.
\newblock 2010.
\newblock Recurrent neural network based language model.
\newblock In {\em INTERSPEECH}, volume~2, page~3.

\bibitem[\protect\citename{Mikolov \bgroup et al.\egroup
  }2013a]{mikolov2013exploiting}
Tomas Mikolov, Quoc~V Le, and Ilya Sutskever.
\newblock 2013a.
\newblock Exploiting similarities among languages for machine translation.
\newblock {\em nternational Conference on Learning Representations (ICLR)}.

\bibitem[\protect\citename{Mikolov \bgroup et al.\egroup
  }2013b]{mikolov2013linguistic}
Tomas Mikolov, Wen-tau Yih, and Geoffrey Zweig.
\newblock 2013b.
\newblock Linguistic regularities in continuous space word representations.
\newblock In {\em HLT-NAACL}, pages 746--751.

\bibitem[\protect\citename{Pennington \bgroup et al.\egroup
  }2014]{pennington2014glove}
Jeffrey Pennington, Richard Socher, and Christopher~D. Manning.
\newblock 2014.
\newblock Glove: Global vectors for word representation.
\newblock In {\em Empirical Methods in Natural Language Processing (EMNLP)},
  pages 1532--1543.

\bibitem[\protect\citename{{Plank} \bgroup et al.\egroup
  }2016]{2016arXiv160405529P}
B.~{Plank}, A.~{S{\o}gaard}, and Y.~{Goldberg}.
\newblock 2016.
\newblock {Multilingual Part-of-Speech Tagging with Bidirectional Long
  Short-Term Memory Models and Auxiliary Loss}.
\newblock {\em ArXiv e-prints}, April.

\bibitem[\protect\citename{Socher \bgroup et al.\egroup
  }2013]{socher2013recursive}
Richard Socher, Alex Perelygin, Jean~Y Wu, Jason Chuang, Christopher~D Manning,
  Andrew~Y Ng, and Christopher Potts.
\newblock 2013.
\newblock Recursive deep models for semantic compositionality over a sentiment
  treebank.
\newblock In {\em Proceedings of the conference on empirical methods in natural
  language processing (EMNLP)}, volume 1631, page 1642. Citeseer.

\bibitem[\protect\citename{West}1996]{douglas1996west}
Douglas~B West.
\newblock 1996.
\newblock Introduction to graph theory.

\end{thebibliography}

\bibliographystyle{acl2016}

\end{document}